\documentclass[sigconf]{acmart}

\usepackage{caption}
\usepackage{subcaption}
\usepackage{framed}
\usepackage{stfloats}
\AtBeginDocument{%
  \providecommand\BibTeX{{%
    \normalfont B\kern-0.5em{\scshape i\kern-0.25em b}\kern-0.8em\TeX}}}

\setcopyright{none}
\renewcommand\footnotetextcopyrightpermission[1]{}
\begin{document}

\title{Fooling State-of-the-Art Deepfake Detection with High-Quality Deepfakes}

\author{Arian Beckmann}
\orcid{0000-0003-4504-3736}
\affiliation{%
  \institution{Fraunhofer Heinrich-Hertz-Institute}
  \city{Berlin}
  \country{Germany}
}
\email{arian.beckmann@hhi.fraunhofer.de}

\author{Anna Hilsmann}
\orcid{0000-0002-2086-0951}
\affiliation{%
  \institution{Fraunhofer Heinrich-Hertz-Institute}
  \city{Berlin}
  \country{Germany}}
\email{anna.hilsmann@hhi.fraunhofer.de}

\author{Peter Eisert}
\orcid{0000-0001-8378-4805}
\affiliation{%
  \institution{Fraunhofer Heinrich-Hertz-Institute}
  \city{Berlin}
  \country{Germany}}
\affiliation{%
  \institution{Humboldt University of Berlin}
  \city{Berlin}
  \country{Germany}}
\email{peter.eisert@hhi.fraunhofer.de}

\renewcommand{\shortauthors}{Arian Beckmann, Anna Hilsmann, \& Peter Eisert}
\begin{abstract}
Due to the rising threat of deepfakes to security and privacy, it is most important to develop robust and reliable detectors. In this paper, we examine the need for high-quality samples in the training datasets of such detectors. Accordingly, we show that deepfake detectors proven to generalize well on multiple research datasets still struggle in real-world scenarios with well-crafted fakes. First, we propose a novel autoencoder for face swapping alongside an advanced face blending technique, which we utilize to generate 90 high-quality deepfakes. Second, we feed those fakes to a state-of-the-art detector, causing its performance to decrease drastically. Moreover, we fine-tune the detector on our fakes and demonstrate that they contain useful clues for the detection of manipulations. Overall, our results provide insights into the generalization of deepfake detectors and suggest that their training datasets should be complemented by high-quality fakes since training on mere research data is insufficient.
\end{abstract}




\maketitle
\pagestyle{plain}

\section{Introduction}
Various deep-learning frameworks enable the manipulation of visual media. Deepfakes denote images and videos that show a face whose identity or expression has been manipulated by a deep neural network. Besides fun gimmicks, deepfakes pose a threat to security and privacy. A well-known misuse of this technology is the creation of fake pornographic content showing people who did not consent to have their data used for this purpose \cite{porn-article}. Another malicious use of deepfakes lies in the impersonation of other identities. One could use a deepfake to bypass security precautions that rely on visual data. Moreover, deepfakes can show influential personalities, such as politicians spreading dangerous misinformation and hence pose a threat to society and democracy \cite{zelensky-article}. To generate a deepfake, deep neural networks, usually convolutional neural networks (CNNs), are utilized to extract visual information and generate novel images. We speak of the person whose face is showing in a deepfake as the "target identity" or "target", while the expression is transferred from the "driving identity" or just "driver". Information extracting CNNs, also called encoders, are used to extract the appearance information from data showing the target person. Depending on the architecture of the deepfake model at hand, the same or another encoder extracts the information on expression, and possibly pose and illumination, from the driver image. The extracted information is then fed to a decoder, usually another CNN, to generate a face image showing the target identity with the expression specified by the driver. Since the appearance of deepfakes, independent developers and researchers have been participating in an adversarial game in which one side tries to improve the visual quality and realism of the fakes while the other aims to detect those robustly. On the one hand, complex model architectures and training procedures, as well as sophisticated extensions to existing models, have been proposed to boost the visual quality of deepfakes \cite{dfl, nirkin2019fsgan, mostgan, hififace, high-res, regionaware, styleface, megaportraits, oneshotmegapixel}. On the other hand, deepfake detectors also employ sophisticated architectures, are extended to consider multi-modal data, and pay attention to (common) artifacts, also those not visible to the human eye. This leads to better detection performance, including increased robustness and generalizability \cite{eyeblinking, headposes, Matern, ffpp, uia-vit, auraldynamics, haliassos2021lips, haliassos2022leveraging}. A bottleneck for the development of efficient detection methods lies in the limited availability of high-quality datasets for training and testing. A handful of deepfake databases are available and regularly used for the training and evaluation of detectors \cite{ffpp, dftimit, Celeb_DF_cvpr20, dfdc, wilddeepfakes}. However, these datasets are subject to various limitations, such as a lack of variability in generation methods. A major weakness of the datasets is the lack of realism of the fakes. Poor visual quality and occasionally used blending procedures lead to visual artifacts that can be detected by the human eye. To generate a large number of fake videos showing a variety of identities, the entire generation process, including the gathering of training data, is usually fully automated. Nevertheless, this can lead to deficient training data, given that a dataset for a single person can include blurry images and even images showing the face of another person. This and the already small amount of training data available lead to poorly trained models, which in turn generate fakes of low visual quality.

In this work, we address the problem of lacking realism in deepfake datasets. We demonstrate that models trained on common benchmarks do not generalize to well-crafted fakes. We create a new set of high-quality deepfakes, which we feed to a state-of-the-art deepfake detector that has been shown to generalize well to unseen datasets and forgery methods. The evaluation shows that the detector struggles with the detection of our fakes, while it performs outstandingly on the detection of their pristine counterparts. We then finetune the detector on our dataset and show that our fakes contain clues for fake detection, thereby highlighting the need for more well-crafted fakes in common deepfake databases. Our contribution is threefold:
\begin{itemize}
    \item We propose a novel architecture for faceswap deepfakes that is simple to train and produces high-quality results.
    \item Additionally, we propose a simple extension to the common blending procedure in faceswaps which leads to fewer blending artifacts.
    \item We provide useful insights on the generalization of deepfake detectors as well as the necessity for well-crafted fakes in their training.
\end{itemize}

\section{Related Works}
\subsection{Deepfake Generation}
A common deepfake approach uses a dual-decoder autoencoder architecture and is trained for two specific identities \cite{faceswap}. With time, a variety of extensions and modifications to this framework were proposed. Overall, newer approaches extract the target appearance from merely a single or a few images in order to obtain generalization with respect to identities.
Several approaches utilize adversarial training to increase the visual quality of the fakes \cite{dfl, nirkin2019fsgan, mostgan, hififace, high-res}. Other works incorporate 3D morphable models \cite{Blanz1999AMM} to encode the expression and pose information of the driver \cite{mostgan, hififace}. Furthermore, methods that manipulate the latent space of a trained StyleGAN \cite{Karras2019stylegan2, high-res, regionaware} were proposed. More recent works aim to generate fakes in higher resolution with strong details, but still lack high perceptive realism \cite{styleface, megaportraits, oneshotmegapixel, high-res}. 
\subsection{Deepfake Detection}
Early detection approaches focus on known clues in deepfakes such as missing eye blinking \cite{eyeblinking} or inconsistent head poses \cite{headposes}. Other works utilize simple CNNs to detect (high-level) visual artifacts caused by the forgery model \cite{Matern, ffpp} or post-processing operations like face blending \cite{facexray}. Due to the increase in the visual quality of fakes, later works focus on detecting low-level artifacts \cite{uia-vit}. A promising line of research aims to detect inconsistencies in motion, especially in movements caused by speaking \cite{auraldynamics, haliassos2021lips, haliassos2022leveraging}, leading to stable performances in cross-manipulation scenarios. 

\subsection{Deepfake Datasets}
The first databases containing deepfake videos were proposed to facilitate the development of deepfake detection algorithms \cite{dftimit, headposes}.
Subsequent works present benchmarks containing more fakes with better visual quality.
A widely used benchmark is the FaceForensics++ (FFPP) dataset \cite{ffpp}, which consists of 1,000 videos collected from YouTube and corresponding fakes generated by six different synthesis methods. The Celeb-DF v2 (CDF) dataset \cite{Celeb_DF_cvpr20} provides over 5,000 fake videos and aims to overcome the issue of low visual quality in deepfake benchmarks. Currently, the largest database is part of the Deepfake Detection Challenge (DFDC) \cite{dfdc}. This dataset consists of more than 100,000 fake sequences generated by eight different synthesis procedures using the recordings of 960 cooperating actors. Recently published works aim to enable the training and testing of deepfake detectors with a focus on generalization to unseen manipulation methods and real-world scenarios \cite{wilddeepfakes, deepfakegamecompetition}. 

\section{Creation of High-Quality Deep Fakes}
This section presents our approach to the generation of deepfakes. Since we want to generate fakes with high visual quality, we use a dual-decoder autoencoder. This architecture, originally proposed by \cite{faceswap}, allows the training of a model for two specific identities and is thus able to learn meaningful and highly detailed representations of their appearances. Once fully trained, the encoder can encode any face image of the two identities into a latent code containing extensive information about the present expression, pose, and illumination. Furthermore, each decoder can transform this code into a face image of its corresponding identity. To further improve deepfake quality, we provide modifications and extensions to the approach in \cite{faceswap}. We use a novel autoencoder utilizing an EfficientNet-B4 \cite{Tan2019EfficientNetRM} as the encoder with several residual blocks in the decoder. Moreover, we propose an advanced blending procedure that produces fewer blending artifacts when merging the forged face of the target with the driver's head. Details on which datasets are used to train our models are given in section \ref{section: experiments}.

The section is structured as follows: First, we describe our data collection process, which results in one training dataset, also called faceset, per identity. Thereafter, we present our autoencoder architecture alongside its training details. The section is concluded with a description of the conversion process, which includes the proposal of our advanced blending procedure. 

\subsection{Faceset Collection} \label{subsection: faceset collection}
\begin{figure}[b]
    \centering
    \includegraphics[width=0.435\textwidth]{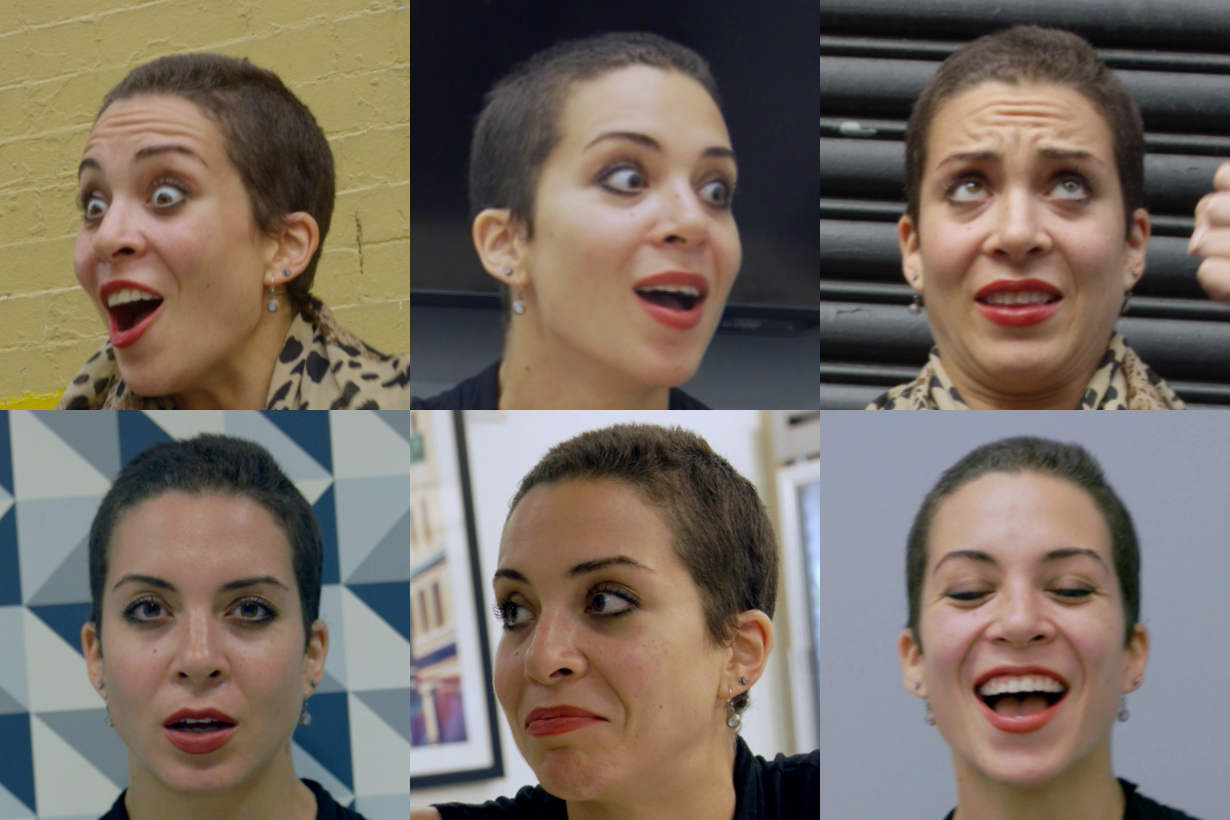}
    \caption{Aligned face images from one of our facesets. Original images are taken from \cite{ffpp}.}
    \label{fig:faceset}
\end{figure}

We now describe the faceset collection procedure, which is adapted to the one in \cite{faceswap}. A faceset is an identity-specific dataset of face images displaying a large variety of expressions, poses, and lighting scenarios. An exemplary subset of one of our facesets is shown in Figure \ref{fig:faceset}.
\begin{figure*}[h]
    \begin{framed}
     \centering
     \begin{subfigure}[b]{0.42\textwidth}
         \centering
         \includegraphics[width=\textwidth]{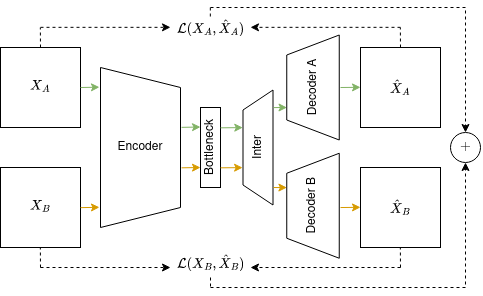}
     \end{subfigure}
    \vrule 
    \begin{subfigure}[b]{0.57\textwidth}
         \centering
         \includegraphics[width=\textwidth]{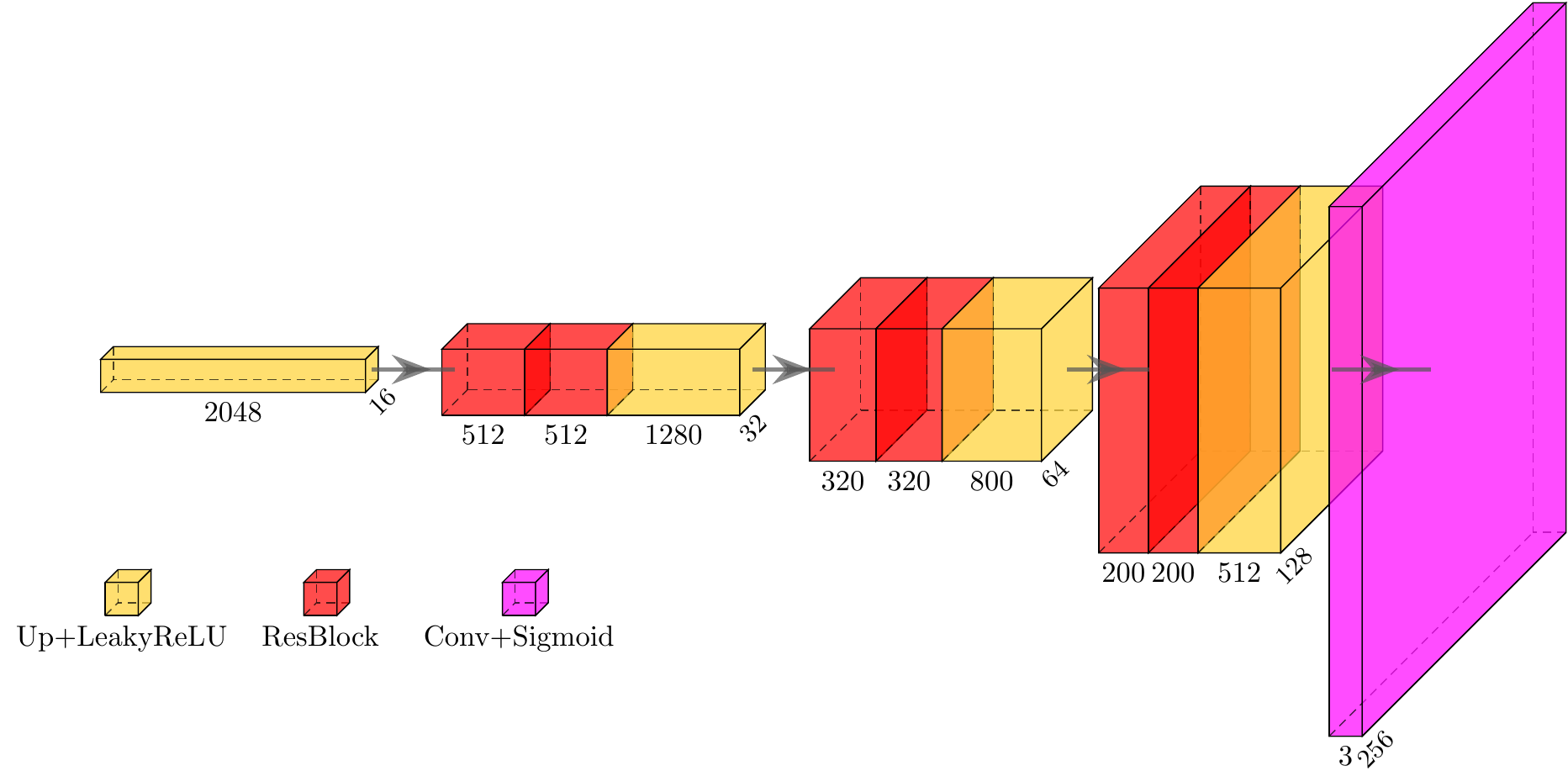}
     \end{subfigure}
     \end{framed}
        \caption{Left: Forward pass of our model. The loss is computed separately for each identity (A and B) and finally summed up. Right: Detailed architecture of the proposed decoder.}
        \label{fig:three graphs}
    \label{fig:autoencoder}
\end{figure*}
In order to train an autoencoder to perform a faceswap between two set identities, we require both their facesets as training data. To collect a faceset, we need to gather multiple videos or images showing the person, possibly from various resources. It is of utmost importance that the data is of high visual quality, as the model will only be able to learn fine visual details if these are visible in the training data. Once the data is gathered, we extract all faces using FAN \cite{fan}. The face detection process also returns $68$ facial landmarks, which are utilized to align the head to a neutral position in the center of the image and crop the image to $512^2$ pixels. Moreover, we use BiSeNet \cite{bisenet-paper, bisnet-github} to obtain a segmentation of the facial regions in each image. This segmentation is essential for the training of the autoencoder, as explained in \ref{subsection: model}. Given the fully extracted faceset, we clean it by removing faces that are blurry, too small (undetailed), in extreme poses, or display a wrong identity. Otherwise, the faceset will contain deficient data that impede the training of the model. To remove images that show a false identity or no face at all, we encode all images into the feature space of a pre-trained VGG Network \cite{vgg}, classify them via $k$-means into $k=25$ clusters, and remove all clusters that do not contain the person of interest. Faces in extreme poses are identified through their yaw and pitch values, which are computed based on the landmarks obtained previously, while small faces are removed based on their face rectangle size. To identify blurry faces, we compute a blur score based on the variance of the Laplacian of the image and remove all images with a score below a threshold which is identified manually for each faceset. Then, we manually scan the faceset for images that were missed by the previous cleaning steps and delete them. Finally, we remove all images that are too similar to each other using dupeguru \cite{dupegure} and ensure that our faceset contains approximately $4,000$-$8,000$ images, which sharply display the person's face in a variety of poses, expressions and illuminations.
\subsection{Model Architecture \& Training} \label{subsection: model}
We employ a dual-decoder autoencoder architecture, proposed by \cite{faceswap}, as our model for deepfake creation. We use the feature extractor of a pre-trained EfficientNet-B4 \cite{Tan2019EfficientNetRM} as the encoder, which is followed by a linear layer, also called the bottleneck. Before the input reaches a decoder, it passes an intermediate block that consists of another linear layer followed by a nearest-neighbor upsampler with LeakyReLU activation. Both decoders in our model share the same architecture, which is inspired by the STOJO model in \cite{faceswap}. We drop their AdaIN block \cite{Karras2019stylegan2}, increase the number of residual blocks per upsample layer to two and utilize a sub-pixel upscaler \cite{subpixel}. Accordingly, we obtain a decoder with four upscale layers that can upscale the output of the intermediate block to a spatial resolution of $256^2$ pixels. The output of each upscaler passes a LeakyReLu activation. Each upscaler, except the last one, is additionally followed by two residual blocks. The final output of the decoder is computed by a single $2D$ convolutional layer with a sigmoid activation. A visualization of our autoencoder, including a detailed representation of the decoder is given in Figure \ref{fig:autoencoder}. 

\subsubsection*{Forward-backward pass}
A single batch (of size one) consists of two input images, one for each identity. We crop the central $80\%$ of the images and resize them to a spatial resolution of $(256, 256)$ before feeding them to the model. A forward pass of these two inputs results in two output images (one per decoder) of size $(256, 256, 3)$. We train the model such that it learns to reconstruct the faces of each identity. For each input-output pair $(X, \hat{X})$ we compute the loss 
\begin{equation*}
\begin{aligned}
    \mathcal{L}(X,\hat{X}) &= \mathcal{L}_{Recon} (X*M_{face}, \hat{X}*M_{face}) \\
    &+ \lambda_{eye} \cdot \mathcal{L}_{Recon}(X*M_{eye}, \hat{X}*M_{eye}) \\
    &+ \lambda_{mouth} \cdot \mathcal{L}_{Recon}(X*M_{mouth}, \hat{X}*M_{mouth}),
\end{aligned}
\end{equation*}
consisting of a reconstruction term, with
\begin{equation*}
    \mathcal{L}_{Recon}(X, \hat{X}) = \mathcal{L}_{DSSIM}(X, \hat{X}) + \mathcal{L}_{MSE}(X, \hat{X}),
\end{equation*}
where $\mathcal{L}_{DSSIM}$ and $\mathcal{L}_{MSE}$ are given by the DSSIM \cite{ssim} and MSE metrics respectively. DSSIM utilizes filters which mimic the sensitivity of the human visual system to changes in (high-)frequency components. The combination of DSSIM and MSE enables our model to achieve both accuracy and visual appeal in reconstructing inputs. Moreover,  the masks $M$ correspond to the respective facial regions denoted in the subscript. The loss terms concerning the eye and mouth are necessary to punish visible artifacts. In line with \cite{faceswap}, we set $\lambda_{eye} = 3$ and $\lambda_{mouth} = 2$. Overall, we are only concerned about the inner part of the face, as we merely want to perform a swap onto another head instead of generating the head entirely. We compute the loss for input-output pairs of both identities at once and then perform an update step on the weights of the model. The forward pass and loss computation are displayed on the left-hand side of Figure \ref{fig:autoencoder}. This training procedure forces the encoder to learn how to encode an image of any of the two identities into a representation of expression, pose, and illumination that is independent of the identity. Furthermore, the decoders learn to transform this latent representation into an image of the respective identity with corresponding attributes.

\paragraph{Training details}
\begin{itemize}
    \item \textit{Optimization.} We use the Adam optimizer \cite{adam} with default $\beta$ and $\epsilon = 1 \times 10^{-7}$. We set the learning rate to $5 \times 10^{-5}$ and train the model for 1,000,000 steps with a batch size of $32$.
    \item \textit{Data Augmentation.} We apply data augmentation to the input and ground truth images. At first, we perform contrast limited adaptive histogram equalization with a chance of $0.5$. Then, the color and lightness parameters in LAB space are randomly adjusted. Moreover, we perform random rotation, scaling, and translation to the color-augmented images. Last, we apply warping to the input images for the first half of training. The warping helps the encoder to generalize its learned representations across identities, while disabling the warping lets the model learn finer details in the end.
\end{itemize}
\subsection{Conversion \& Advanced Blending}
To swap the face of a target identity onto the head of a driver identity, given an appropriately trained model, we extract the driving face from the frame of interest and align it as described in \ref{subsection: faceset collection}. After cropping and resizing, the face image is fed into the model so that the output image is generated by the decoder corresponding to the target identity. This allows us to obtain an image displaying the target with the attributes present in the driver image. Before we re-align the output image and insert it into the driver frame, we apply our advanced blending to merge the output face with the head of the driver. If we want to apply the swap for an entire video, the above procedure is repeated for each frame independently. 

\begin{figure}[t]
    \centering
    \includegraphics[width=0.38\textwidth]{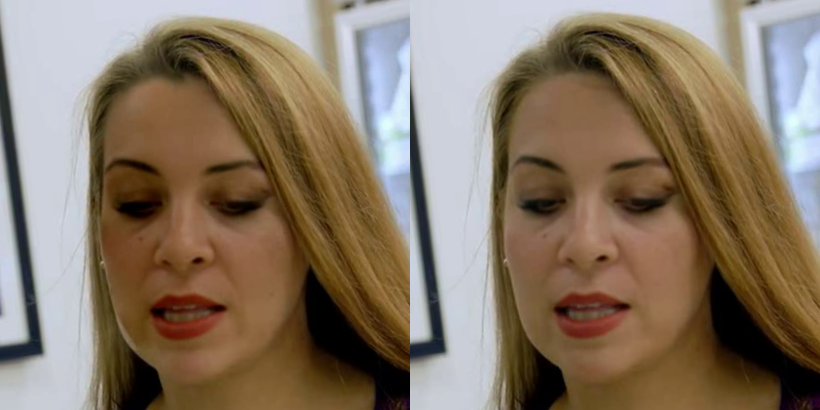}
    \caption{Comparison of the conventional blending procedure and the one proposed by us. Left: Conventional blending with the mask defined by driver. Right: Result with our proposed blending. Best viewed in color.}
    \label{fig:blending}
\end{figure}
\paragraph{Advanced Blending}
The conventional blending procedure utilizes Poisson Blending \cite{poisson-blending} and uses the inner-face segmentation mask of the driver image to indicate where the blending should be performed. However, this can lead to strongly visible artifacts, see Figure \ref{fig:blending}. Artifacts manifest at the blending boundary on the right edge of the face. Additionally, we observe that the lighting around the right eye is inconsistent with the rest of the face. This is caused by the edge of the blending mask being too close to the edge of either of the faces (driver or fake), hence including parts of the image that lie outside of the face, such as hair or background, in the blending process. To reduce these artifacts, we propose a simple but effective adjustment to the blending mask: If the edge of the blending mask is of suitable distance to the edge of the face, the boundary artifacts disappear. Therefore, we squeeze the mask on each side by 15$px$ and thus increase the distance between the edges of the face and the mask. Note that the squeezing amount is variable and should be adapted to both identities for optimal results, but it is kept constant for all our experiments for consistency. The usage of the squeeze mask effectively removes the boundary artifacts. However, occasional inconsistencies in lighting can still appear when an entire video is manipulated. Hence, we compute the face mask of the generated face as well, squeeze it and intersect it with the squeezed mask of the driver to further exclude regions that lie outside of any of the two faces. The resulting mask is finally used for blending. As shown on the right-hand side of Figure \ref{fig:blending}, no artifacts appear when our blending procedure is used.

\begin{figure}[b]
    \centering
    \includegraphics[width=0.435\textwidth]{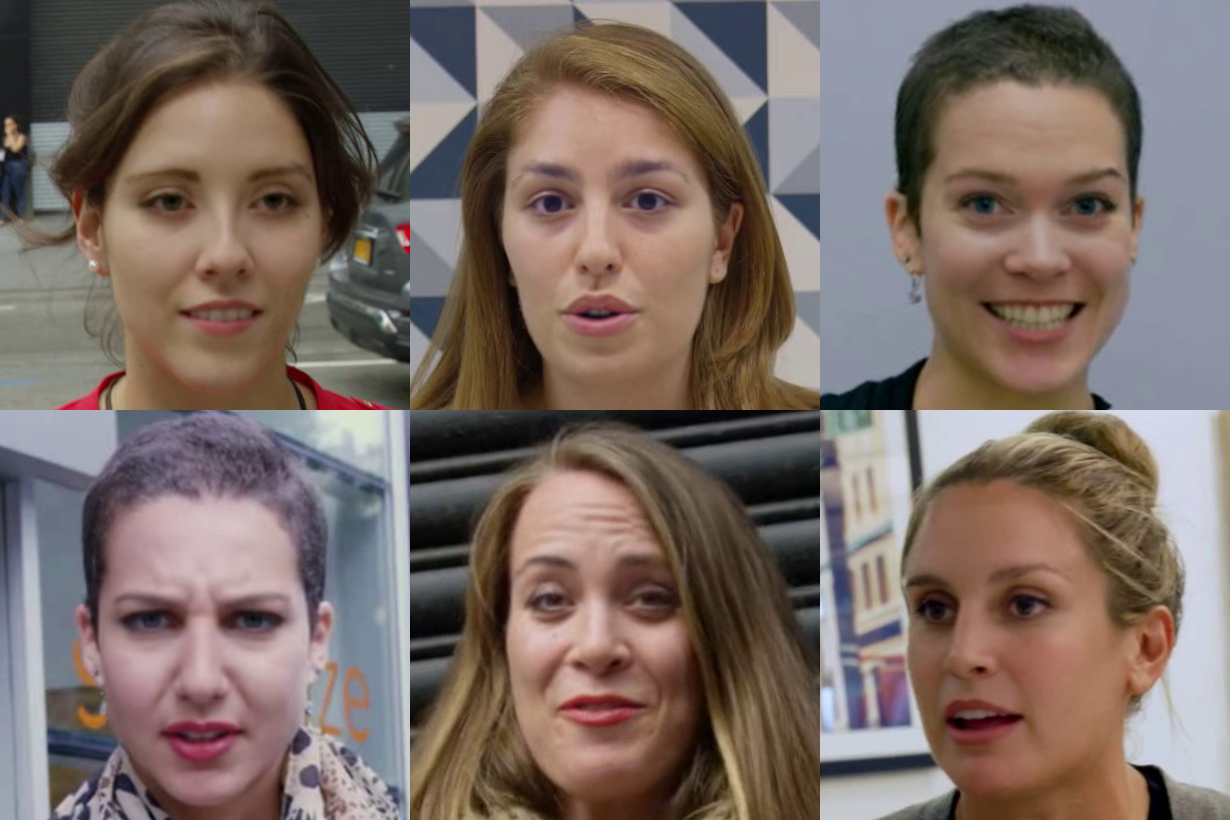}
    \caption{Selected frames of our deepfake dataset. Images generated with six different autoencoders.}
    \label{fig:fakes}
\end{figure}

\section{Experiments} \label{section: experiments}
In this section, we present the experiments conducted in order to demonstrate the realism of our fakes as well as the necessity to include high-quality deepfakes in the training of robust detectors for real-world scenarios. We use the "actors" subset of the deepfake detection dataset \cite{googledeepfakes} to train our deepfake autoencoders. The dataset consists of multiple videos of 28 different actors, displaying a variety of poses, expressions, and head movements. We select 11 of the 28 identities and build 13 identity pairings. A faceset for each identity is obtained as described in \ref{subsection: faceset collection}. We train a model for each identity pairing and swap the faces in selected videos showing the identities corresponding to the respective model obtaining 90 deepfake videos. We use the corresponding driver videos as the pristine counterpart to our set of forged videos. For simplicity, we refer to the union of these data as "our" dataset. The amount of pristine videos is 77, as the same driver video can be used for multiple forgeries. Some qualitative results of our deepfake generation can be seen in Figure \ref{fig:fakes}.

First, we inspect the performance of a state-of-the-art deepfake detector proposed in \cite{haliassos2022leveraging} on our dataset. Thereafter, we use our data to fine-tune the detector and re-evaluate its performance on hold-out testing sets. 

\subsection{Testing On High-Quality Fakes}
The RealForensics detector \cite{haliassos2022leveraging} learns to classify deepfakes by identifying inconsistencies in facial movement instead of looking for merely simple artifacts. This leads to state-of-the-art generalization performance with respect to unseen datasets and forgery methods. Their approach stands out due to a self-supervised learning framework that utilizes the audio of pristine videos in order to help the model learn stronger representations of facial movement. The detector can classify fakes, even when no audio is available for a given video. We download their fully trained model and prepare our data according to their pre-processing. Then, we test the detector on our dataset. The results are reported in Table \ref{tab:testing}. They clearly demonstrate that the detector struggles with the detection of high-quality deepfakes. Moreover, the detector's stellar performance on the pristine videos indicates that the poor performance on the fakes is not caused by the shift to another domain of videos. On the one hand, we conclude that our fakes are of sufficient quality to fool detectors that perform well in cross-dataset and cross-manipulation scenarios. On the other hand, we argue that deepfake detectors which are merely trained on research data struggle with well-crafted fakes in real-world scenarios, despite generalizing well across other research datasets.


\begin{table}
  \caption{Test results of RealForensics on our deepfakes and their pristine counterparts from \cite{ffpp}. Testing on our fakes causes a drop in accuracy from 97.4 to 26.7. All scores in \%.}
  \label{tab:testing}
  \begin{tabular}{cccc}
    \toprule
    AUC& Acc(Pristine)& Acc(Fake)& Acc(Fake+Pristine)\\
    \midrule
    80.2& 97.4& 26.7& 59.3\\
  \bottomrule
\end{tabular}
\end{table}
\begin{table}[t]
  \caption{Average AUC scores (in \%) of 10 finetuned detectors on the testing subsets of Deepfakes(DF), Faceswap(FS), Face2Face(F2F) and NeuralTextures(NT) in FFPP as well CDF.}
  \label{tab:finetune-old}
  \begin{tabular}{ccccc}
    \toprule
    DF & FS & F2F & NT & CDF\\
    \midrule
    98.2& 95.3& 97.7& 97.1 & 78.1\\
  \bottomrule
\end{tabular}
\end{table}
\begin{table*}[b]
  \caption{Test results of the finetuned RealForensics detector on our ten test-splits. All metrics in \%. The Avg.\ metrics are weighted according to the test-split class distributions.}
  \label{tab:finetune-own}
  \begin{tabular}{c|cccccccccc||c}
    & 01 & 02 & 03 & 04 & 05 & 06 & 07 & 08 & 09 & 10 & \textbf{Avg.\ }\\
    \midrule
    AUC & 83.3 & 70.5 & 91.7 & 95.6 & 80.3 & 87.2 & 88.9 & 82.2 & 85.7 & 92.6 & \textbf{86.2} \\
    Acc(Real) & 68.8 & 93.4 & 94.7 & 100 & 70.6 & 61.1 & 94.4 & 100 & 77.8 & 70.6 & \textbf{83} \\
    Acc(Fake) & 83.3 & 50 & 78.9 & 45.8 & 78.9 & 94.1 & 71.4 & 55.5 & 100 & 94.7 & \textbf{75.7} \\
    \# Samples(Real) : \# Samples (Fake) & 16:18 & 16:18 & 19:19 & 17:24 & 17:19 & 18:17 & 18:21 & 15:9 & 18:21 & 17:19 & \textbf{17}:\textbf{19} \\
  \bottomrule
\end{tabular}
\end{table*}
\subsection{Finetuning With High-Quality Fakes}
In order to demonstrate the necessity of high-quality fakes in the training sets of deepfake detectors, we fine-tune the RealForensics detector \cite{haliassos2022leveraging} on our dataset. Given the small size of our dataset, we perform a tenfold cross-validation experiment. We separate the dataset into different splits by randomly sampling two exclusive identities for the test-split and one for the validation-split. For a sampled identity, we gather all videos that show their face either as the target in a fake or as the driver in a pristine video and assign them to the corresponding split. We sample 10 train-, validation- and test-splits and ensure that no duplicate splits appear. Thereafter, we fine-tune the fully trained detector on each split and compute performance metrics on the corresponding test-split.

\paragraph{Training details}
\begin{itemize}
    \item \textit{Optimization.} We use the Adam optimizer \cite{adam} with default $\beta$ and $\epsilon$. We set the learning rate to $8 \times 10^{-5}$ and train both the backbone and the classification head for $50$ epochs with a batch size of $4$.
    \item \textit{Metrics.} We train the model using binary cross-entropy loss, as it is done in \cite{haliassos2022leveraging}. However, we do not use their self-supervised learning approach for the additional training of the backbone. Furthermore, we save the model states with the best performance on the validation-set under the skewed accuracy metric $\lambda_1$Acc(Fake) $+ \lambda_2$Acc(Real). We empirically set $\lambda_1 = 1$ and $\lambda_2 = 3$.
\end{itemize}
Table \ref{tab:finetune-own} shows the accuracy and AUC scores of the fine-tuned detectors on the various test-splits as well as their class distributions. The results clearly indicate that the detector is able to discriminate between our fakes and pristine videos. Hence, we argue that high-quality fakes are able to provide useful information during training. We decided to employ a simple fine-tuning procedure as we aim to demonstrate the utility of the high-quality fakes instead of showing that the detector can perfectly detect our fakes. This can explain the drop in the Acc(Real) metric in Table \ref{tab:finetune-own}. We hypothesize that sophisticated training with high-quality fakes, a larger training corpus, and, potentially, the self-supervised learning approach by \cite{haliassos2022leveraging} can lead to even better and more robust features for the separation of fake and real faces. Furthermore, to ensure that the detectors' original knowledge is not corrupted by the fine-tuning procedure, we evaluate the fine-tuned models on the testing-sets of FFPP \cite{ffpp} and CDF \cite{Celeb_DF_cvpr20}. The results are displayed in Table \ref{tab:finetune-old}. We see that the fine-tuned models still perform well on average on FFPP and CDF. 

\section{Conclusion}
In this paper, we proposed an autoencoder architecture alongside an extension to the blending procedure for deepfake faceswaps. We thoroughly gathered facesets for 11 different identities and used our model and advanced blending to build a dataset containing 90 high-quality deepfakes. We fed our deepfakes to a state-of-the-art deepfake detector, which was shown to generalize well in cross-dataset and cross-manipulation settings. Thereby, we showed that well-performing detectors trained on common research datasets still struggle in real-world scenarios. An experiment in which we used our data to finetune a given detector demonstrated that our high-quality fakes possess additional clues for the detection of fakes. These clues highlight the need for more high-quality fakes in the training process of robust detectors and can potentially lead to better generalization results. The latter argument will be the subject of future work.

\begin{acks}
This work has recieved partial funding by the German Federal Ministry of Education and Research (BMBF) through the Research Program FAKEID under Contract no.\ 13N15735, as well as the Fraunhofer Society in the Max Planck-Fraunhofer collaboration project NeuroHum.
\end{acks}

\bibliographystyle{ACM-Reference-Format}
\bibliography{references}


\end{document}